\pdfoutput=1

\documentclass[9pt]{article}

\usepackage{geometry}
\usepackage{times}  
\usepackage{helvet}  
\usepackage{courier}  
\usepackage{xcolor}
\usepackage[hyphens]{url}  
\usepackage{graphicx} 
\usepackage{booktabs}
\usepackage{enumitem}
\usepackage{amssymb}
\usepackage{natbib}  
\usepackage{caption} 
\DeclareCaptionStyle{ruled}{labelfont=normalfont,labelsep=colon,strut=off} 
\usepackage{algorithm}
\usepackage{algpseudocode}
\usepackage{newfloat}
\usepackage{listings}
\usepackage{amsmath}
\DeclareMathOperator*{\argmax}{argmax} 

\floatstyle{ruled}
\newfloat{listing}{tb}{lst}{}
\floatname{listing}{Listing}

\usepackage{bibentry}

\begin{document}

\title{Quantum-inspired Reinforcement Learning for Synthesizable Drug Design}
\author{
    Dannong Wang\textsuperscript{\rm 1},
    Jintai Chen\textsuperscript{\rm 2}, 
    Yingzhou Lu\textsuperscript{\rm 3}, 
    Minjie Shen\textsuperscript{\rm 4}, 
    Lulu Chen\textsuperscript{\rm 4}, \\
    Zhiding Liang\textsuperscript{\rm 1}, 
    Tianfan Fu\textsuperscript{\rm 1},
    Xiao-Yang Liu\textsuperscript{\rm 1} \\ 
  \textsuperscript{1}Rensselaer Polytechnic Institute (RPI) \\ 
  \textsuperscript{2}University of Illinois Urbana-Champaign (UIUC) \\ 
  \textsuperscript{3}Stanford University \\ 
  \textsuperscript{4}Virginia Tech  \\ 
}

\maketitle

\begin{abstract}
Synthesizable molecular design (also known as synthesizable molecular optimization) is a fundamental problem in drug discovery, and involves designing novel molecular structures to improve their properties according to drug-relevant oracle functions (i.e., objective) while ensuring synthetic feasibility. However, existing methods are mostly based on random search. To address this issue, in this paper, we introduce a novel approach using the reinforcement learning method with quantum-inspired simulated annealing policy neural network to navigate the vast discrete space of chemical structures intelligently. Specifically, we employ a deterministic REINFORCE algorithm using policy neural networks to output transitional probability to guide state transitions and local search using genetic algorithm to refine solutions to a local optimum within each iteration. Our methods are evaluated with the Practical Molecular Optimization (PMO) benchmark framework with a 10K query budget. We further showcase the competitive performance of our method by comparing it against the state-of-the-art genetic algorithms-based method. 
\end{abstract}


\section{Introduction}

Novel types of safe and effective drugs are needed to meet the medical needs of billions worldwide and improve the quality of human life. 
The process of discovering a new drug candidate and developing it into an approved drug for clinical use is known as \textit{drug discovery and development}. Two distinct stages in the process are: 
\begin{itemize}
\item {\it Drug discovery} focuses on identifying novel drug molecules with desirable pharmaceutical properties; 
\item {\it Drug development} aims to test the drug's safety and efficacy in human bodies via clinical trials~\cite{chen2024trialbench}. 
After the clinical trials, the results are reviewed by the US Food and Drug Administration (FDA) or equivalent government bodies from other countries. 
Upon approval, the new drug will be available for clinical use.
\end{itemize}

Drug discovery and development is notoriously time-consuming, labor-intensive, and expensive. 
Bringing a novel drug to the market currently takes 13-15 years and requires 2-3 billion US dollars on average~\cite{lu2024uncertainty}. 
Efficient and safe drug discovery has garnered growing interest, especially after the worldwide COVID-19 pandemic~\cite{wu2022cosbin}. 
Artificial Intelligence (AI) and Machine Learning (ML) are the latest attempts to make this process more efficient and accurate with the help of machine learning models trained on a large amount of historical data~\cite{chen2024uncertainty}.

Synthesizable molecular design is a fundamental task in drug discovery, aiming to enhance the desirable properties of molecules while ensuring they remain synthetically feasible. Specifically, the task involves optimizing a molecular structure with respect to an oracle function~\cite{synnet}. This process often involves navigating a very large discrete space, where traditional methods can be computationally expensive and limited in their exploration capabilities~\cite{gao2022samples}. SynNet is a synthesis-based library that uses neural networks to probabilistically model the synthetic trees and applies a Genetic Algorithm (GA) to manipulate binary fingerprints that represent molecules, and it shows success in this task~\cite{synnet}.

Recently, reinforcement learning (RL) algorithms, specifically the deterministic REINFORCE (dREINFORCE) algorithm~\cite{}, has demonstrated success in solving challenging combinatorial optimization problems, including the graph max-cut problem and the Ising Spin Glasses Model problem~\cite{}. This type of deterministic policy gradient algorithm, which samples trajectories and updates the policy using computed gradients from rewards, has shown promising results in these domains~\cite{l2a,lu2022cot}.

Inspired by these successes, this paper investigates the application of quantum-inspired dREINFORCE method to the molecular optimization problem by replacing the GA in SynNet in an attempt to see improved results. 
The main contributions of this paper can be summarized as follows:
\begin{itemize}[leftmargin=*]
 \item \textbf{Method}. We propose a quantum-inspired reinforcement learning method  specifically designed for synthesizable drug molecular design.
 \item We show the first successful attempt that formulates the synthesizable drug design as an intelligent discrete search problem defined on molecular fingerprint space, which suppress random-walk behavior and explore the chemical space intelligently. 
\item \textbf{Results}. We validate the effectiveness of the proposed method across 23 pharmaceutical property optimization tasks in practical molecular optimization (PMO) benchmark and find that our method achieves competitive performance compared with a couple of cutting-edge molecular design methods. 
\end{itemize}



\section{Related Works}
\paragraph{AI-driven Molecular Design Methods.}
Molecular generation techniques present a promising approach for the automated design of molecules with specific pharmaceutical properties, such as synthetic accessibility and drug-likeness. These methods can be broadly categorized based on their approach to generating or searching for molecules: (1) deep generative models (DGMs), which emulate the distribution of molecular data, including variational autoencoders (VAE)~\cite{gomez2018automatic,jin2018junction}, generative adversarial networks (GAN)~\cite{guimaraes2017objective,cao2018molgan}, normalizing flow models~\cite{shi2019graphaf,luo2021graphdf}, and energy-based models~\cite{liu2021graphebm,fu2022antibody}; and (2) combinatorial optimization methods that directly search within the discrete chemical space, encompassing genetic algorithms (GA)~\cite{jensen2019graph,nigam2019augmenting,gao2021amortized}, reinforcement learning (RL) approaches~\cite{Olivecrona,You2018-xh,zhou2019optimization,jin2020multi,glass2021moler,ahn2020guiding,fu2022reinforced}, Bayesian optimization (BO)~\cite{korovina2020chembo}, Markov Chain Monte Carlo (MCMC)~\cite{fu2021mimosa,bengio2021gflownet}, and gradient ascent~\cite{fu2021differentiable, shen2021deep}.

\paragraph{Quantum-inspired Methods. }

\section{Methodology}

\begin{figure*}
 \centering
 \includegraphics[width=0.92\linewidth]{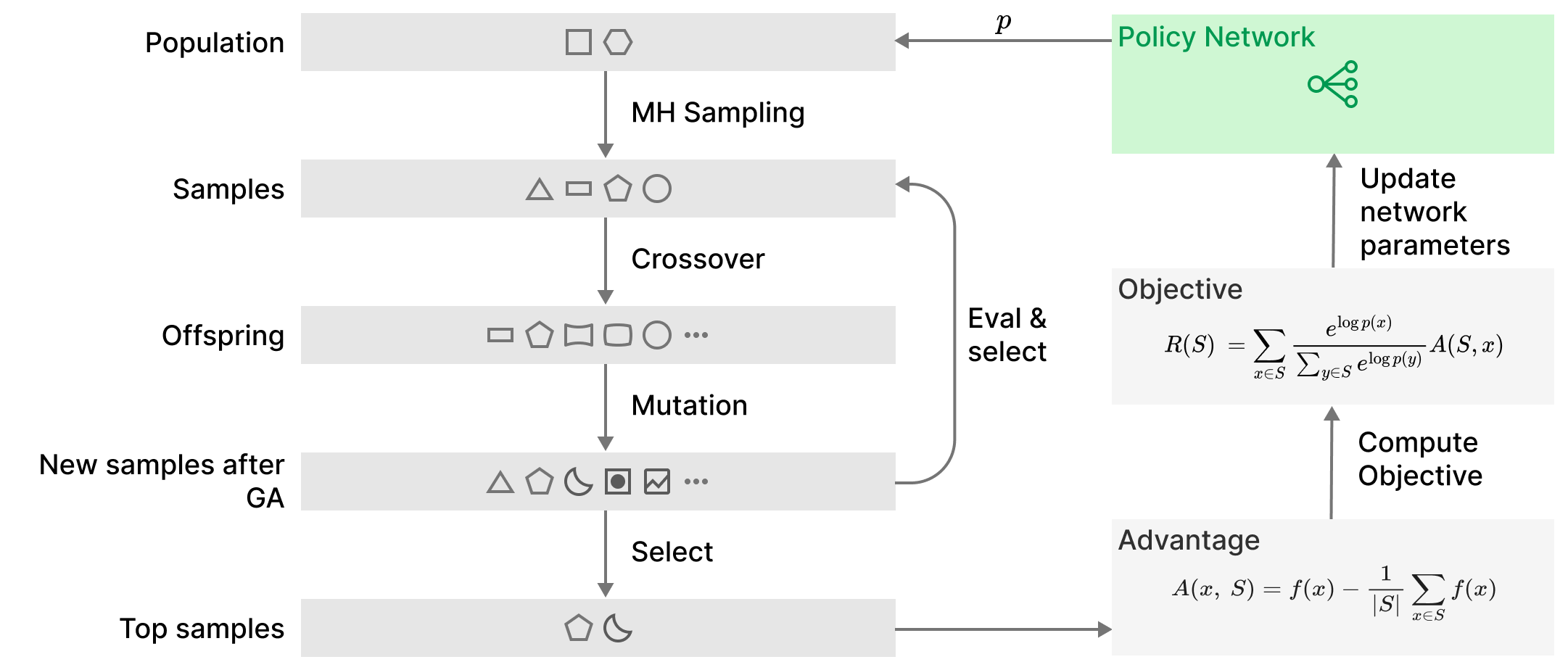}
 \caption{Framework of quantum-inspired dREINFORCE method. It conducts iterative search. In each iteration, it leverages policy network to select the sampling probabilities for the bits in molecular fingerprint.  }
 \label{fig:framework}
\end{figure*}

\subsection{Problem: Molecular Design} 
The molecular design problem can be formulated as the following discrete optimization problem:
\[m^* = \argmax_{m \in \mathcal{M}}\ \mathcal{O}(m), \]
where $m$ represents the molecular structure, $\mathcal{M}$ represents the whole chemical space that contains all valid molecules (around $10^{60}$~\cite{bohacek1996art}), and $\mathcal{O}: \mathcal{M} \xrightarrow[]{} \mathbb{R}$ represents the oracle function, which evaluates the properties of the molecules and returns a scalar. The oracle is considered a black box. In realistic drug discovery, it could be a high-fidelity molecular simulation process, e.g., molecular docking, and takes intensive computational resources. Due to the high cost of oracles, it is necessary to limit the number of oracle calls to a certain budget~\cite{gao2022samples}. 

\subsection{Quantum-inspired Reinforcement Learning}

\begin{algorithm}
\caption{Monte Carlo Policy Gradient \cite{chen2023montecarlopolicygradient}}\label{alg:cap}
\begin{algorithmic}
\Require number of population $k$, oracle function $\mathcal{O}$, budget $b$, number of samples per molecules after sampling $m$
\State Randomly select $k$ molecules $s_1, ..., s_k$ from database;
\State Initialize the policy network (probabilistic model) with parameters $\theta$;

\While{ number of oracle calls within budget}

    \State Get $p$ from policy network;
    \State $S = \{s_1^1, ..., s_1^m\}, ..., \{s_k^1, ..., s_k^m\} \gets \mathtt{mh\_sampling}(s_i,...,s_k, p, m)$ 
    \State $S, V \gets \mathtt{local\_search}(S, \mathcal{O})$ \Comment{Local search using GA, which also returns $V$, the fitness of $S$}
    \State $\hat{s_1}, ..., \hat{s_k} \gets \mathtt{select\_top}(S, V)$ \Comment{Select top $k$ molecules with best fitness score}
    
    \State $P \gets p.\mathtt{repeats}(k \cdot m)$ \Comment{Repeat the probability distribution for all molecules}
    \State $P^S \gets \mathtt{where}(S, P, 1 - P)$ \Comment{Probability of S (if $S_{i,j}$ = 0, $P^S_{i,j} = 1 - P^S_{i,j}$, else $P^S_{i,j} =  P^S_{i,j}$ )}
    \State $P'^{S} \gets \sum_i^m \log(P^S) $ \Comment{Log probability}
    \State Compute advantage $A$, where $(A_{i,j} = V_{i,j} - V_{i}\mathtt{.mean}()) / V_i.\mathtt{std()} $
    \State Compute objective $o = \sum_i^m \sum_j^k  \mathtt{softmax}(P'^{S})A_{i,j}$
    \State Update parameter $\theta$ using gradient descent from $o$

\EndWhile
\end{algorithmic}
\end{algorithm}


For ease of exposition, we illustrate our method in Figure~\ref{fig:framework}. 
This approach is inspired by quantum annealing~\cite{qa}, which can be used to find a global optimum over a large search space. While classical simulated annealing relies on a classical temperature parameter to control exploration and exploitation~\cite{delahaye2019simulated}, we rely on the learning dynamics of the neural network policy. Initially, the untrained policy network will propose transitions to states with lower fitness. As the network learns and the policy improves, it increasingly suggests transitions to states with higher fitness. The behavior is similar to simulated annealing, where initially, higher temperatures prioritize exploration, and later, lower temperatures prioritize exploitation~\cite{qa}. 

We begin by sampling a random population from the initial dataset of molecules. The population is represented by binary Morgan fingerprints~\cite{synnet}. For each iteration, we obtain probabilities by performing a forward pass through the policy network. Then, we sample the next state based on these probabilities, perform a local search, compute the reward and the policy gradient, and finally update the network parameters.

\textbf{Environment}: The synthetic tree decoder and the oracle functions are part of the deterministic environment. Given a binary fingerprint, it returns the corresponding score, which serves as the reward.

\textbf{Policy Network}: The network consists of a single layer with trainable parameters, taking no input and outputting transition probabilities corresponding to the number of bits in the Morgan fingerprints. 

\textbf{Sampling}: We use Metropolis-Hastings sampling that uses probabilities from the policy network to guide exploration. By flipping a limited number of bits in molecular fingerprints, we balance the need for exploration while making sure the molecular structural validity.

\textbf{Local Search}: We adapt the SynNet~\cite{synnet} genetic algorithm as a local search strategy, employing a reduced number of iterations for computational efficiency. This refinement step optimizes sampled candidates toward local optima.

\section{Experiments}
In this section, we discuss the experimental results. We start by describing the experimental setup and implementation details, then demonstrate the experimental results and analyze the results. 

\subsection{Experimental Setup}
We follow the Practical Molecular Optimization (PMO) benchmark~\cite{gao2022samples} to set up the experiment.
We establish SynNet GA~\cite{synnet} as a baseline and compare the performance metrics against our dREINFORCE method.

\noindent\textbf{Oracle:} We select DRD2~\cite{drd2}, GSK3$\beta$~\cite{chen2021data}, JNK3~\cite{gsk3b,lu2019integrated}, and QED~\cite{qed} as pharmaceutical-related oracle functions. They are implemented by the Therapeutic Data Commons (TDC)~\cite{tdc} library. The first three objectives are machine learning models that predict the response of molecules against these proteins: dopamine receptor type 2, c-Jun N-terminal kinases-3, and
glycogen synthase kinase 3$\beta$~\cite{synnet}. QED (Quantitative Estimate of Druglikeness) measures the druggability of molecules. All oracle scores are normalized from 0 to 1, where 1 is optimal~\cite{gao2022samples}.

\noindent\textbf{Evaluation Metrics:} We report top-1, top-10, and top-100 average, as well as the top-10, and top-100 Area Under the Curve (AUC), and Synthetic Accessibility (SA) and top-100 diversity as metrics and limit the number of oracle calls to 10000 to ensure practicality. The top-$K$ metrics show the average and the standard deviation of the top-$K$ molecules generated. The top-$K$ AUC metrics, designed in~\cite{gao2022samples}, show average value versus the number of Oracle calls. It rewards methods that reach high averages using fewer Oracle calls.  The Synthetic Accessibility (SA) score measures how hard it is to synthesize a given chemical compound. The SA score ranges from 0 to 10, a lower SA score means the molecule is easier to synthesize and is more desirable. 
Top-100 diversity measures the averaged internal distance within the top-100 molecules~\cite{gao2022samples}. 
{Diversity} of generated molecules is defined as the average pairwise Tanimoto distance between the Morgan fingerprints~\citep{synnet,fu2021mimosa}. 
$\text{diversity} = 1 - \frac{1}{|\mathcal{Z}|(|\mathcal{Z}|-1)}\sum_{Z_1,Z_2 \in \mathcal{Z}, Z_1 \neq Z_2} \text{sim}(Z_1,Z_2),$ where $\mathcal{Z}$ is the set of generated molecules. $\text{sim}(Z_1,Z_2)$ is the Tanimoto similarity between molecule $Z_1$ and $Z_2$. 
{(Tanimoto) Similarity} measures the similarity between the input molecule and generated molecules. 
It is defined as
$\text{sim}(X,Y) = \frac{\mathbf{b}_{X}^\top \mathbf{b}_{Y}}{\Vert \mathbf{b}_{X}\Vert_2 \Vert \mathbf{b}_{Y}\Vert_2}$,  
$\mathbf{b}_{X}$ is the binary Morgan fingerprint vector for the molecule $X$. 

\noindent\textbf{Data:} We randomly sample the initial population from the ZINC 250K dataset~\cite{zinc}. 
The ZINC database is a comprehensive, freely available resource that contains commercially available compounds for virtual screening and drug discovery research. It is specifically designed to help researchers identify potential drug candidates by providing a curated collection of ``drug-like'' molecules. ZINC includes chemical structures in ready-to-dock formats, enabling seamless integration into computational drug design workflows. We also used a selection of random seeds to sample and ensure generalizability. 

\noindent\textbf{Molecular representations: } We represent molecules using Morgan fingerprints with length 4096 and radius 2 for both the baseline SynNet GA algorithm and our dREINFORCE algorithm.

\subsection{Implementation Details}

\noindent\textbf{SynNet GA}: We use the genetic algorithm from SynNet~\cite{synnet} as a baseline. The initial population size is 16, the size of the offspring is 64, the probability of mutation is 0.5, and the number of mutations per element is 24.  

\noindent\textbf{dREINFORCE}: We also use an initial population size of 16. Each trajectory is repeated 8 times after the Metropolis-Hastings sampling algorithm and run through 6 iterations of local search using GA with the off-spring size of 256, a mutation probability of 0.5. The policy neural network is a single-layer neural network. It takes no explicit input and has one output layer with a dimension of 4096 and sigmoid activation. 
It is initialized with random values between 0.49 and 0.51. Adam optimizer is used with learning rate 1e-3.

\begin{table*}[h!]
\centering
\caption{Performance comparison between SynNet GA and Our method based on \textbf{Average Top-1} ($\uparrow$) from 5 independent runs.}
\begin{tabular}{l|cc}
\toprule
\textbf{Oracle} & SynNet GA & dREINFORCE\\
\midrule
DRD2 &  0.990 $\pm$ 0.013 & 0.988 $\pm$ 0.016 \\
GSK3$\beta$ &  0.816 $\pm$ 0.103 & 0.812 $\pm$ 0.055 \\
JNK3 &  0.542 $\pm$ 0.085 &  0.696 $\pm$ 0.032 \\
QED & 0.948 $\pm$ 0.000& 0.947 $\pm$ 0.001 \\
albuterol\_similarity & {0.723 $\pm$ 0.154} & 0.644 $\pm$ 0.094 \\
amlodipine\_mpo & {0.615 $\pm$ 0.018} & 0.604 $\pm$ 0.022 \\
Aripiprazole\_Similarity & {0.816 $\pm$ 0.065} & 0.796 $\pm$ 0.051 \\
Celecoxib\_Rediscovery & 0.478 $\pm$ 0.027 & {0.486 $\pm$ 0.048} \\
deco\_hop & 0.616 $\pm$ 0.008 & {0.663 $\pm$ 0.098} \\
isomers\_c7h8n2o2 & {0.981 $\pm$ 0.038} & 0.976 $\pm$ 0.047 \\
isomers\_c9h10n2o2pf2cl & {0.812 $\pm$ 0.091} & 0.793 $\pm$ 0.102 \\
median1 & 0.286 $\pm$ 0.058 & {0.278 $\pm$ 0.029} \\
median2 & 0.323 $\pm$ 0.183 & {0.349 $\pm$ 0.209} \\
mestranol\_similarity & {0.527 $\pm$ 0.076} & 0.495 $\pm$ 0.139 \\
Osimertinib\_mpo & 0.804 $\pm$ 0.019 & {0.816 $\pm$ 0.009} \\
perindopril\_mpo & {0.575 $\pm$ 0.018} & 0.558 $\pm$ 0.017 \\
ranolazine\_mpo & 0.799 $\pm$ 0.012 & {0.803 $\pm$ 0.016} \\
scaffold\_hop & 0.502 $\pm$ 0.020 & {0.522 $\pm$ 0.015} \\
sitagliptin\_mpo & {0.401 $\pm$ 0.065} & 0.369 $\pm$ 0.026 \\
thiothixene\_rediscovery & 0.382 $\pm$ 0.033 & {0.404 $\pm$ 0.049} \\
troglitazone\_rediscovery & 0.307 $\pm$ 0.040 & {0.321 $\pm$ 0.033} \\
Valsartan\_SMARTS & 0.144 $\pm$ 0.288 & {0.157 $\pm$ 0.315} \\
zaleplon\_mpo & 0.503 $\pm$ 0.025 & {0.504 $\pm$ 0.017} \\
\bottomrule
\end{tabular}
\label{tab:average_top1}
\end{table*}

\begin{table*}[h!]
\centering
\caption{Performance comparison between SynNet GA and Our method based on \textbf{Average Top-10} ($\uparrow$) from 5 independent runs.}
\begin{tabular}{l|cc}
\toprule
\textbf{Oracle} & SynNet GA & dREINFORCE\\
\midrule
DRD2 &  0.981 $\pm$ 0.019 & 0.952 $\pm$ 0.050 \\
GSK3$\beta$ & 0.779 $\pm$ 0.094 & 0.777 $\pm$ 0.047 \\
JNK3 &  0.481 $\pm$ 0.077 & 0.666 $\pm$ 0.037 \\
QED &  0.946 $\pm$ 0.001& 0.944 $\pm$ 0.004 \\
albuterol\_similarity & {0.613 $\pm$ 0.121} & 0.568 $\pm$ 0.062 \\
amlodipine\_mpo & 0.588 $\pm$ 0.026 & {0.579 $\pm$ 0.016} \\
Aripiprazole\_Similarity & {0.781 $\pm$ 0.060} & 0.755 $\pm$ 0.047 \\
Celecoxib\_Rediscovery & 0.436 $\pm$ 0.023 & {0.439 $\pm$ 0.049} \\
deco\_hop & 0.608 $\pm$ 0.006 & {0.639 $\pm$ 0.060} \\
Isomers\_c7h8n2o2 & {0.907 $\pm$ 0.043} & 0.901 $\pm$ 0.031 \\
isomers\_c9h10n2o2pf2cl & {0.729 $\pm$ 0.071} & 0.693 $\pm$ 0.080 \\
median1 & {0.242 $\pm$ 0.024} & 0.232 $\pm$ 0.014 \\
median2 & 0.271 $\pm$ 0.109 & {0.302 $\pm$ 0.147} \\
mestranol\_similarity & {0.478 $\pm$ 0.076} & 0.442 $\pm$ 0.079 \\
Osimertinib\_mpo & 0.784 $\pm$ 0.018 & {0.806 $\pm$ 0.007} \\
perindopril\_mpo & {0.552 $\pm$ 0.026} & 0.539 $\pm$ 0.012 \\
ranolazine\_mpo & 0.785 $\pm$ 0.010 & {0.794 $\pm$ 0.010} \\
scaffold\_hop & 0.490 $\pm$ 0.013 & {0.509 $\pm$ 0.014} \\
sitagliptin\_mpo & {0.316 $\pm$ 0.040} & 0.308 $\pm$ 0.016 \\
thiothixene\_rediscovery & 0.361 $\pm$ 0.027 & {0.375 $\pm$ 0.056} \\
troglitazone\_rediscovery & {0.288 $\pm$ 0.036} & 0.285 $\pm$ 0.019 \\
Valsartan\_SMARTS & 0.131 $\pm$ 0.262 & {0.149 $\pm$ 0.298} \\
zaleplon\_mpo & 0.477 $\pm$ 0.024 & {0.478 $\pm$ 0.016} \\
\bottomrule
\end{tabular}
\label{tab:average_top10}
\end{table*}

\begin{table*}[h!]
\centering
\caption{Performance comparison between SynNet GA and Our method based on \textbf{Average Top-100} ($\uparrow$) from 5 independent runs.}
\begin{tabular}{l|cc}
\toprule
\textbf{Oracle} & SynNet GA & dREINFORCE\\
\midrule
DRD2 & 0.897 $\pm$ 0.103 & 0.795 $\pm$ 0.202 \\
GSK3$\beta$ & 0.650 $\pm$ 0.112 &  0.673 $\pm$ 0.049 \\
JNK3 &  0.383 $\pm$ 0.075 & 0.610 $\pm$ 0.033 \\
QED &  0.935 $\pm$ 0.006& 0.930 $\pm$ 0.010 \\
albuterol\_similarity & 0.511 $\pm$ 0.083 & 0.476 $\pm$ 0.037 \\
amlodipine\_mpo & 0.543 $\pm$ 0.009 & 0.550 $\pm$ 0.014 \\
Aripiprazole\_Similarity & 0.704 $\pm$ 0.066 & 0.672 $\pm$ 0.035 \\
Celecoxib\_Rediscovery & 0.379 $\pm$ 0.029 & 0.377 $\pm$ 0.040 \\
deco\_hop & 0.594 $\pm$ 0.004 & 0.606 $\pm$ 0.010 \\
Isomers\_c7h8n2o2 & 0.697 $\pm$ 0.105 & 0.672 $\pm$ 0.060 \\
isomers\_c9h10n2o2pf2cl & 0.580 $\pm$ 0.056 & 0.525 $\pm$ 0.060 \\
median1 & 0.200 $\pm$ 0.011 & 0.184 $\pm$ 0.011 \\
median2 & 0.196 $\pm$ 0.014 & 0.230 $\pm$ 0.046 \\
mestranol\_similarity & 0.408 $\pm$ 0.071 & 0.363 $\pm$ 0.012 \\
Osimertinib\_mpo & 0.751 $\pm$ 0.021 & 0.784 $\pm$ 0.008 \\
perindopril\_mpo & 0.513 $\pm$ 0.026 & 0.505 $\pm$ 0.010 \\
ranolazine\_mpo & 0.757 $\pm$ 0.011 & 0.773 $\pm$ 0.012 \\
scaffold\_hop & 0.472 $\pm$ 0.010 & 0.497 $\pm$ 0.013 \\
sitagliptin\_mpo & 0.203 $\pm$ 0.037 & 0.204 $\pm$ 0.022 \\
thiothixene\_rediscovery & 0.329 $\pm$ 0.023 & 0.336 $\pm$ 0.048 \\
troglitazone\_rediscovery & 0.261 $\pm$ 0.030 & 0.250 $\pm$ 0.010 \\
Valsartan\_SMARTS & 0.040 $\pm$ 0.081 & 0.123 $\pm$ 0.246 \\
zaleplon\_mpo & 0.429 $\pm$ 0.021 & 0.433 $\pm$ 0.009 \\
\bottomrule
\end{tabular}
\label{tab:average_top100}
\end{table*}

\begin{table*}[h!]
\centering
\caption{Performance comparison between SynNet GA and Our method based on \textbf{AUC Top-10} ($\uparrow$) from 5 independent runs.}
\begin{tabular}{l|cc}
\toprule
\textbf{Oracle} & SynNet GA & dREINFORCE\\
\midrule
DRD2 &  0.926 $\pm$ 0.040 & 0.859 $\pm$ 0.081 \\
GSK3$\beta$ &  0.704 $\pm$ 0.084 & 0.678 $\pm$ 0.024 \\
JNK3 &  0.390 $\pm$ 0.059 & 0.511 $\pm$ 0.060 \\
QED &  0.922 $\pm$ 0.002& 0.915 $\pm$ 0.006 \\
albuterol\_similarity & 0.568 $\pm$ 0.098 & 0.532 $\pm$ 0.052 \\
amlodipine\_mpo & 0.559 $\pm$ 0.020 & 0.550 $\pm$ 0.015 \\
Aripiprazole\_Similarity & 0.741 $\pm$ 0.057 & 0.705 $\pm$ 0.044 \\
Celecoxib\_Rediscovery & 0.411 $\pm$ 0.011 & 0.406 $\pm$ 0.048 \\
deco\_hop & 0.591 $\pm$ 0.005 & 0.610 $\pm$ 0.038 \\
Isomers\_c7h8n2o2 & 0.833 $\pm$ 0.037 & 0.834 $\pm$ 0.041 \\
isomers\_c9h10n2o2pf2cl & 0.649 $\pm$ 0.064 & 0.633 $\pm$ 0.082 \\
median1 & 0.228 $\pm$ 0.022 & 0.207 $\pm$ 0.009 \\
median2 & 0.263 $\pm$ 0.106 & 0.287 $\pm$ 0.147 \\
mestranol\_similarity & 0.444 $\pm$ 0.051 & 0.409 $\pm$ 0.088 \\
Osimertinib\_mpo & 0.760 $\pm$ 0.017 & 0.771 $\pm$ 0.006 \\
perindopril\_mpo & 0.529 $\pm$ 0.027 & 0.505 $\pm$ 0.015 \\
ranolazine\_mpo & 0.752 $\pm$ 0.013 & 0.745 $\pm$ 0.017 \\
scaffold\_hop & 0.475 $\pm$ 0.011 & 0.487 $\pm$ 0.010 \\
sitagliptin\_mpo & 0.272 $\pm$ 0.039 & 0.268 $\pm$ 0.019 \\
thiothixene\_rediscovery & 0.343 $\pm$ 0.023 & 0.337 $\pm$ 0.050 \\
troglitazone\_rediscovery & 0.277 $\pm$ 0.034 & 0.267 $\pm$ 0.019 \\
Valsartan\_SMARTS & 0.128 $\pm$ 0.255 & 0.145 $\pm$ 0.291 \\
zaleplon\_mpo & 0.454 $\pm$ 0.026 & 0.452 $\pm$ 0.015 \\
\bottomrule
\end{tabular}
\label{tab:auc_top10}
\end{table*}

\begin{table*}[h!]
\centering
\caption{Performance comparison between SynNet GA and Our method based on \textbf{AUC Top-100} ($\uparrow$) from 5 independent runs.}
\begin{tabular}{l|cc}
\toprule
\textbf{Oracle} & SynNet GA & dREINFORCE\\
\midrule
albuterol\_similarity & 0.471 $\pm$ 0.072 & 0.438 $\pm$ 0.032 \\
amlodipine\_mpo & 0.517 $\pm$ 0.013 & 0.518 $\pm$ 0.013 \\
Aripiprazole\_Similarity & 0.655 $\pm$ 0.053 & 0.620 $\pm$ 0.033 \\
Celecoxib\_Rediscovery & 0.354 $\pm$ 0.020 & 0.347 $\pm$ 0.037 \\
deco\_hop & 0.578 $\pm$ 0.004 & 0.581 $\pm$ 0.003 \\
Isomers\_c7h8n2o2 & 0.567 $\pm$ 0.106 & 0.545 $\pm$ 0.080 \\
isomers\_c9h10n2o2pf2cl & 0.473 $\pm$ 0.047 & 0.435 $\pm$ 0.079 \\
median1 & 0.188 $\pm$ 0.012 & 0.166 $\pm$ 0.011 \\
median2 & 0.189 $\pm$ 0.013 & 0.213 $\pm$ 0.048 \\
mestranol\_similarity & 0.374 $\pm$ 0.052 & 0.332 $\pm$ 0.016 \\
Osimertinib\_mpo & 0.723 $\pm$ 0.023 & 0.736 $\pm$ 0.007 \\
perindopril\_mpo & 0.487 $\pm$ 0.026 & 0.464 $\pm$ 0.014 \\
ranolazine\_mpo & 0.715 $\pm$ 0.018 & 0.707 $\pm$ 0.022 \\
scaffold\_hop & 0.456 $\pm$ 0.010 & 0.471 $\pm$ 0.007 \\
sitagliptin\_mpo & 0.156 $\pm$ 0.038 & 0.152 $\pm$ 0.021 \\
thiothixene\_rediscovery & 0.310 $\pm$ 0.020 & 0.300 $\pm$ 0.042 \\
troglitazone\_rediscovery & 0.250 $\pm$ 0.027 & 0.233 $\pm$ 0.011 \\
Valsartan\_SMARTS & 0.039 $\pm$ 0.079 & 0.120 $\pm$ 0.240 \\
zaleplon\_mpo & 0.397 $\pm$ 0.026 & 0.379 $\pm$ 0.012 \\
\bottomrule
\end{tabular}
\label{tab:auc_top100}
\end{table*}

\begin{table*}[h!]
\centering
\caption{Performance comparison between SynNet GA and Our method based on \textbf{diversity} ($\uparrow$) from 5 independent runs.}
\begin{tabular}{l|cc}
\toprule
\textbf{Oracle} & SynNet GA & dREINFORCE\\
\midrule
DRD2 &  0.711 $\pm$ 0.060 &  0.744 $\pm$ 0.051 \\
GSK3$\beta$ & 0.682 $\pm$ 0.102 & 0.617 $\pm$ 0.136 \\
JNK3 & 0.728 $\pm$ 0.066 & 0.526 $\pm$ 0.013 \\
QED & 0.754 $\pm$ 0.020 & 0.783 $\pm$ 0.041 \\
albuterol\_similarity & 0.778 $\pm$ 0.032 & 0.810 $\pm$ 0.012 \\
amlodipine\_mpo & 0.734 $\pm$ 0.029 & 0.649 $\pm$ 0.074 \\
Aripiprazole\_Similarity & 0.678 $\pm$ 0.042 & 0.659 $\pm$ 0.071 \\
Celecoxib\_Rediscovery & 0.685 $\pm$ 0.064 & 0.722 $\pm$ 0.069 \\
deco\_hop & 0.780 $\pm$ 0.027 & 0.774 $\pm$ 0.016 \\
Isomers\_c7h8n2o2 & 0.808 $\pm$ 0.028 & 0.798 $\pm$ 0.033 \\
isomers\_c9h10n2o2pf2cl & 0.807 $\pm$ 0.053 & 0.825 $\pm$ 0.021 \\
median1 & 0.720 $\pm$ 0.094 & 0.795 $\pm$ 0.017 \\
median2 & 0.757 $\pm$ 0.078 & 0.725 $\pm$ 0.065 \\
mestranol\_similarity & 0.700 $\pm$ 0.103 & 0.762 $\pm$ 0.035 \\
Osimertinib\_mpo & 0.790 $\pm$ 0.016 & 0.731 $\pm$ 0.043 \\
perindopril\_mpo & 0.712 $\pm$ 0.036 & 0.699 $\pm$ 0.060 \\
ranolazine\_mpo & 0.718 $\pm$ 0.048 & 0.633 $\pm$ 0.095 \\
scaffold\_hop & 0.780 $\pm$ 0.025 & 0.759 $\pm$ 0.018 \\
sitagliptin\_mpo & 0.799 $\pm$ 0.020 & 0.820 $\pm$ 0.014 \\
thiothixene\_rediscovery & 0.697 $\pm$ 0.042 & 0.640 $\pm$ 0.114 \\
troglitazone\_rediscovery & 0.746 $\pm$ 0.037 & 0.745 $\pm$ 0.053 \\
Valsartan\_SMARTS & 0.825 $\pm$ 0.019 & 0.840 $\pm$ 0.014 \\
zaleplon\_mpo & 0.734 $\pm$ 0.066 & 0.756 $\pm$ 0.064 \\
\bottomrule
\end{tabular}
\label{tab:diversity}
\end{table*}

\begin{table*}[h!]
\centering
\caption{Performance comparison between SynNet GA and Our method based on \textbf{Synthetic Accessibility (SA)} ($\downarrow$) from 5 independent runs.}
\begin{tabular}{l|cc}
\toprule
\textbf{Oracle} & SynNet GA & dREINFORCE\\
\midrule
DRD2 &   2.851 $\pm$ 0.145 & 3.173 $\pm$ 0.155 \\
GSK3$\beta$ &  3.471 $\pm$ 0.458 & 4.301 $\pm$ 0.453 \\
JNK3 &  3.941 $\pm$ 0.272 & 4.158 $\pm$ 0.494 \\
QED &   2.883 $\pm$ 0.233 & 2.848 $\pm$ 0.140 \\
albuterol\_similarity & 2.810 $\pm$ 0.157 & 2.594 $\pm$ 0.098 \\
amlodipine\_mpo & 2.972 $\pm$ 0.230 & 2.885 $\pm$ 0.272 \\
Aripiprazole\_Similarity & 2.407 $\pm$ 0.299 & 2.420 $\pm$ 0.223 \\
Celecoxib\_Rediscovery & 2.528 $\pm$ 0.125 & 2.683 $\pm$ 0.282 \\
deco\_hop & 3.352 $\pm$ 0.180 & 3.277 $\pm$ 0.089 \\
Isomers\_c7h8n2o2 & 2.423 $\pm$ 0.213 & 2.273 $\pm$ 0.069 \\
isomers\_c9h10n2o2pf2cl & 2.694 $\pm$ 0.266 & 2.448 $\pm$ 0.191 \\
median1 & 3.516 $\pm$ 0.254 & 3.618 $\pm$ 0.118 \\
median2 & 3.006 $\pm$ 0.272 & 2.945 $\pm$ 0.115 \\
mestranol\_similarity & 3.185 $\pm$ 0.176 & 3.339 $\pm$ 0.271 \\
Osimertinib\_mpo & 3.369 $\pm$ 0.417 & 3.345 $\pm$ 0.206 \\
perindopril\_mpo & 3.705 $\pm$ 0.259 & 4.048 $\pm$ 0.250 \\
ranolazine\_mpo & 3.895 $\pm$ 0.482 & 3.847 $\pm$ 0.148 \\
scaffold\_hop & 3.322 $\pm$ 0.156 & 3.365 $\pm$ 0.133 \\
sitagliptin\_mpo & 2.807 $\pm$ 0.104 & 3.151 $\pm$ 0.148 \\
thiothixene\_rediscovery & 2.535 $\pm$ 0.141 & 2.932 $\pm$ 0.478 \\
troglitazone\_rediscovery & 3.285 $\pm$ 0.269 & 3.505 $\pm$ 0.279 \\
Valsartan\_SMARTS & 2.910 $\pm$ 0.226 & 2.991 $\pm$ 0.279 \\
valsartan\_smarts & 2.848 $\pm$ 0.197 & 3.139 $\pm$ 0.303 \\
zaleplon\_mpo & 2.646 $\pm$ 0.222 & 2.755 $\pm$ 0.148 \\
\bottomrule
\end{tabular}
\label{tab:sa}
\end{table*}

\section{Results \& Analysis}
For each optimization property, we conduct 5 independent runs with different random seeds to provide a more reliable assessment of the algorithm's performance. 
The results are reported in Table~\ref{tab:average_top1}, \ref{tab:average_top10}, \ref{tab:average_top100}, \ref{tab:auc_top10}, \ref{tab:auc_top100}, \ref{tab:diversity} and~\ref{tab:sa}. 
While performing similarly in most oracles, dREINFORCE outperforms SynNet GA in some tasks. These results demonstrate the potential of reinforcement learning in the drug design task to suppress random-walk behavior of traditional genetic algorithm.


\section{Conclusion}

In this paper, we introduced a novel application of the dREINFORCE algorithm for synthesizable molecular design, aimed at improving drug discovery outcomes. By integrating quantum-inspired reinforcement learning with a neural network-driven policy, we effectively addressed the challenges of navigating the complex chemical space. Our extensive evaluation, conducted using the PMO molecular design benchmark, demonstrated that our method offers competitive performance compared to traditional genetic algorithm approaches. The promising results underscore the potential of quantum-inspired methods in advancing the field of drug discovery, particularly in optimizing molecular properties while ensuring synthetic accessibility. Future work will focus on further refining this approach and exploring its application to broader molecular design tasks.

\bibliographystyle{unsrt}
\bibliography{reference}

\end{document}